\documentclass{article}

\usepackage{amssymb,amsmath}

\usepackage{arxiv}

\usepackage[utf8]{inputenc} 
\usepackage[T1]{fontenc}    
\usepackage{hyperref}       
\usepackage{url}            
\usepackage{booktabs}       
\usepackage{amsfonts}       
\usepackage{nicefrac}       
\usepackage{microtype}      
\usepackage{cleveref}       
\usepackage{lipsum}         
\usepackage{graphicx}
\usepackage{natbib}
\usepackage{doi}

\usepackage{lineno,hyperref}

\usepackage{mathtools,esint}
\usepackage{algpseudocode}
\usepackage{algorithmicx}
\usepackage{algorithm}
\usepackage{afterpage}

\title{Using Deep Reinforcement Learning to solve Optimal Power Flow problem with generator failures}

\author{ \href{https://orcid.org/0000-0002-4082-8385}{\includegraphics[scale=0.06]{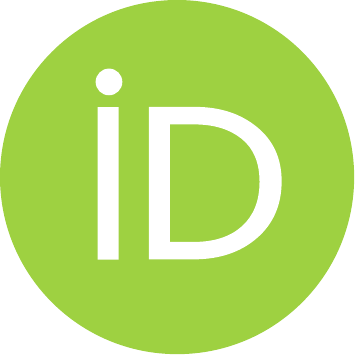}\hspace{1mm}Muhammad Usman Awais}\thanks{} \\
	Department of Computer Science\\
	National University of Computer and Emerging Science\\
	Lahore, Pakistan \\
	\texttt{usman.awais@nu.edu.pk} \\
}

\renewcommand{\shorttitle}{\textit{arXiv} Template}

\hypersetup{
	pdftitle={A template for the arxiv style},
	pdfsubject={q-bio.NC, q-bio.QM},
	pdfauthor={David S.~Hippocampus, Elias D.~Striatum},
	pdfkeywords={First keyword, Second keyword, More},
}
	
\begin{document}
\maketitle

\newcommand{\Env}{\mathcal{E}}
\newcommand{\ObsT}[1]{\mathbf{x}_{t#1}}
\newcommand{\Stat}[1]{\mathbf{s}_{#1}}
\newcommand{\StatEl}[1]{s_{#1}}
\newcommand{\StatS}{\mathcal{S}}
\newcommand{\Act}[1]{\mathbf{a}_{#1}}
\newcommand{\ActS}{\mathcal{A}}
\newcommand{\Rew}[1]{r_{#1}}
\newcommand{\Reward}{R}
\newcommand{\GradTheta}[1]{\nabla_{\theta^{#1}} }
\newcommand{\PolTheta}[1]{\pi_{\theta^{#1}}}
\newcommand{\QPi}{Q^{\pi}}
\newcommand{\QMu}{Q^{\mu}}
\newcommand{\ThetQ}{\theta^{Q}}
\newcommand{\ThetMu}{\theta^{\mu}}
\newcommand{\PolGradTheta}{\GradTheta{} J (\PolTheta{}) }
\newcommand{\Critic}[1]{Q(\Stat{#1},\Act{}|\ThetQ)}
\newcommand{\Actor}[1]{\mu(\Stat{#1}|\ThetMu)}
\newcommand{\Space}[2]{\mathbf{#1_{\mathcal{#2}}}}
\newcommand{\quotes}[1]{``#1''}
\newcommand\norm[1]{\left\lVert#1\right\rVert}
\newcommand{\SimFMU}{\mathbb{F}}
\graphicspath{{./figures/}}
\newcommand{\Ext}{.pdf}
\newcommand{\addPlot}[4]{\begin{figure}[h] \centering \includegraphics[width=#1\textwidth]{#2} \caption{#3} \label{#4} \end{figure}}

\newcommand{\addTwoPlots}[5]{
	\begin{figure}[h]
		\centering
		\begin{tabular}{cc}
			\hspace{-0.6cm}
			\includegraphics[width=#1\textwidth]{#2}&
			\includegraphics[width=#1\textwidth]{#3}\\
		\end{tabular}
		\vspace{-0.6cm}
		\caption{#4}
		\label{#5}
	\end{figure}
}

\newcommand{\addThreePlots}[6]{
	\begin{figure}[h]
		\centering
		\begin{tabular}{ccc}
			\includegraphics[width=#1\textwidth]{#2}&
			\includegraphics[width=#1\textwidth]{#3}&
			\includegraphics[width=#1\textwidth]{#4}\\
		\end{tabular}
		\caption{#5}
		\label{#6} 
	\end{figure}
}

\begin{abstract}
Deep Reinforcement Learning (DRL) is being used in many domains. One of the biggest advantages of DRL is that it enables the continuous improvement of a learning agent. Secondly, the DRL framework is robust and flexible enough to be applicable to problems of varying nature and domain. Presented work is evidence of using the DRL technique to solve an Optimal Power Flow (OPF) problem. Two classical algorithms have been presented to solve the OPF problem. The drawbacks of the vanilla DRL application are discussed, and an algorithm is suggested to improve the performance. Secondly, a reward function for the OPF problem is presented that enables the solution of inherent issues in DRL. Reasons for divergence and degeneration in DRL are discussed, and the correct strategy to deal with them with respect to OPF is presented.    
\end{abstract}


\section{Introduction}
Besides its environmental impact, fossil fuels are not infinite. The world has to move towards a more stable energy source as some geographical areas are already witnessing the depletion of fossil fuels. According to \cite{martins2019analysis} by 2050, the European reserves of fossil fuels may deplete to 14\% of the original. Solar and wind energy are two of the most promising energy sources, but their instability is one of their severe drawbacks. There are different sources of their instability including, variation in weather, internal equipment fault, or an external device failure. Renewable energy sources mentioned above are deployed in a distributed manner according to their geographical suitability. Their physical distance adds to the complexity of system stability. A modern controller in a power grid needs to take action in case of a fault while considering the complete power grid and its distributed generators as one entity, as compared to the traditional local remedies. Due to their distributed nature, the renewable sources must be accompanied by equipment that enables the controller to take unified action on the grid in case of load or generation variation or in case of a fault.

Despite their unstable nature, there are measures devised to make solar and wind energy sources more predictable, controllable, and resilient to change. In order to make the system controllable, modern power systems are multi-sourced renewable energy systems. They are a collection of solar farms, wind turbines, storage batteries, and fossil fuel generators. The last three components play a key role in the stability of electrical supply \cite{guerin2011hybrid}. An operator has to do a tricky job to assign parameters to these components such that the complete grid keeps working in an optimal fashion. Formally, the problem of finding the control parameters to operate the power grid at nominal values is called the Optimal Power Flow(OPF) problem. In the presented work, a methodology is proposed to find control variables using Reinforcement Learning (RL) such that a near-optimal power flow can be achieved in a power grid in case of generator failures.

\section{Related Work}
\label{sec:relatedwork}

Due to the importance of stability of electrical supply, there has been significant research on devising measures of automatic control of grid appliances. Many have used RL for this purpose. In \cite{ernst2004power} authors have laid out the foundations of using RL with power girds. Although at that time, the DRL domain was not mature enough, so they had to rely on the discretization of states to use the traditional RL methods for their solutions. In \cite{dimeas2010multi} and \cite{li2012optimal} authors have used state-based reward functions and discrete actions to train multiple agents. Again a traditional RL method is used, a discretization method is used in the absence of a pure continuous scheme. Recently, in \cite{diao2019autonomous} authors have employed DRL for optimal control of IEEE 14 bus system. They have proposed an agent which they call as \textbf{Grid Mind}. The agent applies the proposed actions on a simulation and checks for any boundary violations that may occur as a result of that action. If there are no violations, the agent observes the final state based on which the reward is calculated. Here again, a discrete approach was used for training the agent. In \cite{huang2019adaptive} authors have proposed a dynamic simulation environment to train DRL agents. InterPSS power system simulator has been used. The environment is used to find values of discrete variables, for example, when to apply generator brakes and use transformers to develop a coordinated under-voltage load shedding scheme. DQN algorithm was used to train a  control policy for applying the load shedding. In \cite{yang2019two} authors have proposed a mixed approach using physics-based optimization and DRL for short-term and long-term optimization. As the problem is formulated in terms of switching the shunt capacitors on or off so the actions are discrete, and there is no need to use a continuous DRL method. In \cite{hua2019optimal} authors have proposed a control system trained using the DRL for Energy Internet (EI). In \cite{wang2019safe} authors have proposed a new algorithm called Constrained Soft Actor-Critic algorithm for volt-var control. In \cite{wu2020battery} authors have applied the Deterministic Policy Gradient Algorithm to optimize the power allocation of a hybrid electric bus. L2RPN challenge \cite{marot2020learning} is an example of promoting DRL in the domain of the smart grid. The aim of L2RPN is to use DRL for control strategies, specifically after eliminating the underlying complexities of the complete model of power grid for speed and performance.

In the presented article, the DRL is used to search for the optimal reference voltages of a power grid in case of generator failures. Here, a complete mathematical model is used to simulate the Power Grid and calculate load flow values. The model is derived from the OpenIPSL library that was prepared after years of research \cite{baudette2018openipsl}. As the model is closer to the physical appliances, so the results are more reliable and applicable to real-world situations.

In the next sections, first the OPF problem is formulated as an RL problem in section \ref{sec:problem}. In section \ref{sec:casestudies}, case studies are discussed. In section \ref{sec:limits}, limitations of the proposed simple solution are discussed. Then in section \ref{sec:AlgoParallel} and \ref{sec:rewardFunc}, two improvements over the basic algorithm are discussed. In section \ref{sec:discussion}, a brief discussion and future research directions are presented. At the end, in section \ref{sec:conc} the conclusive remarks end the article.

\section{Formulating OPF as RL Problem}
\label{sec:problem}

\subsection{Optimal Power Flow Problem}
\label{sec:OptimalPF}
The Optimal Power Flow (OPF) problem has been studied in-depth, yet it still has many aspects to be discovered \cite{mohagheghi2018survey}. It is a highly non-convex problem whose solutions have been proposed using many different techniques including, quadratic programming, linear programming, Lagrange relaxation, interior-point methods, evolutionary computing, fussy logic, and neural networks \cite{lavaei2011zero}. The OPF problem has a mixture of continuous and discrete variables and has an element of uncertainty involved. It can be categorized as a stochastic mixed-integer nonlinear programming problem \cite{mohagheghi2018survey}. However, in the presented work, the OPF is studied from a dynamical system point of view using a Modelica-based simulation component, so there are no stochastic elements considered. Such simplifications are commonly adopted in the literature \cite{mohagheghi2018survey}.

The formal specification of the OPF problem is described in different ways depending on the solution to be proposed. Here the formulation is chosen to identify its dynamical system view as well as keeping intact the essence of how optimization problems are specified in the literature. Equation \ref{eq:OPFOpt} describes the power grid in terms of a dynamical system. $\Space{X}{}$ is the vector space of all continuous state variables. $\Space{I}{}$ is the vector space of input variables. Here, $\Space{P}{}$ represents parameters whose specific values take the system to the optimal state. 

\begin{equation}
\label{eq:OPFOpt}
\begin{array} {lcl} \arg\min\limits_{\mathbf{p}} &&  y(\mathbf{x}(t),\mathbf{u}(t),\mathbf{p})  \quad t \in [0,\infty)\\   
					where \\
					&& \dot{\mathbf{x}(t)} = f(\mathbf{x}(t), \mathbf{u}(t), \mathbf{p}) \\
					&& \mathbf{x} \in \Space{X}{} , \quad  \mathbf{u} \in \Space{I}{}, \quad \mathbf{p} \in \Space{P}{}
\end{array}
\end{equation}		

In equation \ref{eq:constraints} are enumerated the constrains of the optimization problem. Here at every place, $P$ represents the active power, while $Q$ represents the reactive power. $S$ represents the apparent power. $V$ either represents the voltage of a bus ($V_k$) or the voltage between the two points of a line ($V_{\mathcal{L}}$). $\theta$ represents the voltage angle of a bus.  All of these values are represented as continuous state variables inside the model.

\begin{equation}
\label{eq:constraints}
\begin{array} {lcl}	&& \mathcal{N} = {1,2,3, \dots , n} \quad \text{(Buses in the system)} \\
					&& \mathcal{G} \subseteq \mathcal{N} \quad \text{(Generators in the system)}\\
					&& \mathcal{L} \subseteq \mathcal{N} \times \mathcal{N} \quad \text{(Lines in the system)}\\
					&& P_{k}^{min} \le P_{G_k} \le P_{k}^{max}, \quad \forall k \in \mathcal{G} \\
					&& Q_{k}^{min} \le Q_{G_k} \le Q_{k}^{max}, \quad \forall k \in \mathcal{G} \\
					&& V_{k}^{min} \le |V_{k}| \le V_{k}^{max}, \quad \forall k \in \mathcal{N} \\
					&& \theta_{k}^{min} \le |\theta_{k}| \le \theta_{k}^{max}, \quad \forall k \in \mathcal{N} \\
					&& |S_{lm}| \le S_{lm}^{max}, \quad \forall (l,m) \in \mathcal{L} \\
					&& |P_{lm}| \le P_{lm}^{max}, \quad \forall (l,m) \in \mathcal{L} \\
					&& |V_{l} - V_{m}| \le \Delta V_{lm}^{max}, \quad \forall (l,m) \in \mathcal{L} \\
					&& \Space{P}{G} = \{P_k | k \in \mathcal{G}\} \\
					&& \Space{Q}{G} = \{Q_k | k \in \mathcal{G}\} \\
					&& \Space{V}{N} = \{V_k | k \in \mathcal{N}\} \\
					&& \Space{\Theta}{N} = \{\theta_k | k \in \mathcal{N}\} \\
					&& \Space{S}{L} = \{S_{(l,m)} | (l,m) \in \mathcal{L}\} \\
					&& \Space{P}{L} = \{P_{(l,m)} | (l,m) \in \mathcal{L}\} \\
					&& \Space{V}{L} = \{V_{(l,m)} | (l,m) \in \mathcal{L}\} \\
					&& \{ \Space{P}{G} \cup  \Space{Q}{G} \cup \Space{V}{N} \cup \Space{\Theta}{N} \cup \Space{S}{L} \cup \Space{P}{L} \cup \Space{V}{L}\} \subset \Space{X}{}\\
\end{array}
\end{equation}

There are equality constraints for power generation and the load on the system. Apparent power can be categorized into three types: $S_{Gi}$ generated power flowing into the bus $i$, $S_{Li}$ demand power flowing out of the bus $i$, and  $S_{Ti}$ transmitted power flowing through the network to the adjacent buses and the equipment. Equation \ref{eq:apparent} shows the relationship among these three types.

\begin{equation}
	\label{eq:apparent}
	S_{Gi} = S_{Li} + S_{Ti}
\end{equation}

Similarly, because active power $P$ and reactive power $Q$ are components of apparent power $S$ so the same holds for them too \cite{wood2013power}, i.e.
\begin{equation}
	\label{eq:pq}
	\begin{array} {lcl}
		&& P_{Gi} = P_{Li} + P_{Ti} \\
		&& Q_{Gi} = Q_{Li} + Q_{Ti} \\
	\end{array}	
\end{equation}

Equation \ref{eq:eqconstraints} shows the active and reactive power flow equations. If $Y_{ij}$ is the admittance magnitude between the bus $i$ and $j$, then $G_{ij}$ represents the real part of the admittance while $B_{ij}$ represents the imaginary part, and $Y_{ij} = G_{ij} + B_{ij}$ \cite{wood2013power}.  

\begin{equation}
	\label{eq:eqconstraints}
	\begin{array} {lcl}
		&& P_i = \sum\limits_{j \in \mathcal{N}}  ( V_i V_j G_{ij} \cos \theta_{ij} + V_i V_j B_{ij} \sin \theta_{ij}) \\
		&& Q_i = - \sum\limits_{j \in \mathcal{N}}  ( V_i V_j B_{ij} \cos \theta_{ij} + V_i V_j G_{ij} \sin \theta_{ij}) \\
	\end{array}	
\end{equation}

The topological constraints in power grids originate from Kirchoff's voltage law and Kirchoff's current law. As the Modelica models are written directly using a mathematical language, Modelica provides the natural representation of all types of equality constraints. In the presented work, they are part of the Modelica model provided by the OpenIPSL \cite{lin2018novel}. Any parametric value of the system conflicting with the equality constraints will lead to a degenerate solution. 

\subsection{Reinforcement Learning}
\label{sec:RL}

As mentioned earlier, the OPF problem is solved using Reinforcement Learning (RL), so it is appropriate to have a basic introduction of the RL framework.

Traditionally the reinforcement learning framework is described by an agent working on the environment $\Env$ through its actions $\Act{t}$, at a certain time step $t$. The set of all actions is named $\ActS$. The $\Env$ is observable through the observation vector $\ObsT{}$. At each time step, the agent wins a reward $\Rew{t}$ as a result of performing an action $\Act{t}$. As the problem domain suggests, all the actions are real-valued, so $\Act{t} \in \mathbb{R}^n$ holds. Considering the environment $\Env$ to be a dynamical system, the state of the system at time $t$ is $\Stat{t} \in \mathbb{R}^m$, and the set of all states is named $\StatS$. Assuming that the agent's behavior is represented by a policy $\pi$. The probability of selecting an action $\Act{}$ at a state $\Stat{}$ is induced by a policy $\pi$, that is $\pi : \StatS{} \rightarrow P(\ActS)$.

At each time $t$ an action $\Act{t}$ performed on the system takes the system to a new state $\Stat{t}$ from $\Stat{t-1}$. In the present discussion, it is assumed that the system is fully observable so $\Stat{t} = \ObsT{}$. If the Markovian assumption is not applied, the probability distribution of a state is conditioned over all previous states and actions $p(\Stat{t}|\Stat{t-1},\Act{t-1},\dots,\Stat{1},\Act{1})$. Instead, it is assumed that the initial distribution $p(\Stat{1})$ is given and the system at each state $\Stat{t}$ holds Markov's property, making it independent of the complete trajectory of previous states.
	
At each step, the agent is being rewarded by a reward function  $R:S\times A \times S \rightarrow \mathbb{R}$. Its value is defined to be the discounted sum of future rewards $R_t = \sum\limits_{i=t}^T \gamma^{(i-t)} r (\Stat{i},\Act{i}) $, where discounting factor $\gamma \in [0,1]$. A distribution representing the total sum of future rewards at a state is the state's \textit{state-value}.

\paragraph{Action value}The value of policy $\pi$ is defined to be the total discounted reward $V^{\pi}(\Stat{t}) = \mathbb{E}[R|\Stat{t},\pi]$. 
As the system is supposed to entail the Markov property so another similar function $Q^{\pi}(\Stat{t},\Act{t}) = \mathbb{E}[R|\Stat{t},\Act{t},\pi]$ is defined that is called the action-value function. At a specific state, it assigns the value to an action. Practically, it means to take the action $\Act{t}$ at the state $\Stat{t}$ and then following the policy $\pi$ thereafter. The action-value can be viewed as a distribution of rewards obtained by taking a certain action in the pool of experiences.

\paragraph{Performance Objective} For a learning problem, there must be a performance objective that is improved over time. To define that, the probability density of $\Stat{t}$ after $t$ transitions from state $\Stat{1}$ to $\Stat{t}$ is represented by $p(\Stat{1} \rightarrow \Stat{t}, t, \pi)$. The discounted probability distribution of such a state is represented by 
\[\rho^{\pi}(\Stat{t}) \coloneqq \int_{\StatS} \sum_{t=1}^{\infty} \gamma^{t-1} p(\Stat{1}) p(\Stat{1} \rightarrow \Stat{t}, t, \pi) ds\]. Using these terms the performance objective can be represented as follows

\begin{equation}
\label{eq:performObj}
\begin{array} {lcl} J(\pi_{\theta})  & = & \int_{\StatS} \rho^{\pi}(\Stat{}) \int_{\ActS} \pi_{\theta} (\Stat{},\Act{})  r(\Stat{},\Act{}) da \, ds  \\ & = & \mathbb{E}_{s \sim \rho^{\pi}, a \sim \pi_{\theta}} [r(s,a)] \end{array}
\end{equation}

Maximizing the above performance objective will give the value of actions $\Act{}$ for which the system has optimal values \cite{Silver2014}. 

\paragraph{Policy Gradient} One major improvement in reinforcement learning algorithms is devised by the policy gradient theorem. Improving the parameter $\theta$ in the direction of policy gradient $\PolGradTheta{}$ enables the search algorithm to converge faster. Computing the value of $\PolGradTheta$ is reduced to a simple expectation \cite{sutton2000policy}.

\begin{equation}
\label{eq:polGrad}
\begin{array} {lcl} \PolGradTheta  & = & \int_{\StatS} \rho^{\pi}(\Stat{}) \int_{\ActS} \GradTheta{} \PolTheta{} (\Act{}|\Stat{}) \QPi (\Stat{},\Act{}) da \, ds  \\ & = & \mathbb{E}_{s \sim \rho^{\pi}, a \sim \pi_{\theta}} [\GradTheta{} \log \PolTheta{} (\Act{}|\Stat{}) \QPi (\Stat{},\Act{})] \end{array}
\end{equation}

The above conditions only apply when the actions and states are continuous variables, which is true for the case under consideration.  

\paragraph{Deep Reinforcement Learning}

As mentioned in section \ref{sec:RL}, \textit{state-value} and  \textit{action-value} are important parts of reinforcement learning. When tabular methods of storing actions, state, state-values, and action-values are not feasible. Due to a large number of states or actions. It is preferred to use approximate methods to represent \textit{action-value} and \textit{state-value} functions. When deep neural networks are used for such approximations, then it is called Deep Reinforcement Learning (DRL). Typically in equations \ref{eq:performObj} and \ref{eq:polGrad}, $\theta$ is considered as the weights of the neural network, which needs to be improved so that the policy $\pi$ behaves similar to the hidden real policy of the model.

\subsection{Formulation}
\label{sec:formulation}

The OPF problem can be compared to a multi-dimensional multi-arm bandit problem. There are a few differences and a few similarities. In the multi-arm bandit problem, the initial state of the system is always the same. After applying an action $\Act{}$, the system goes through an observable change that defines the reward earned by $\Act{}$. There are some important differences from the bandit setting. One, the OPF problem is continuous in both actions and states. Second, the action space is multi-dimensional and infinite, and there are some constraints over its values as mentioned in equation \ref{eq:constraints}. 

\subsubsection{Elements of RL}

\paragraph{Constraints} In the presented work, the constraints are simplified to keep the complexity manageable.  The primary focus of the study is voltage regulation of the system, i.e. $V_{k}^{min} \le |V_{k}| \le V_{k}^{max}, \quad \forall k \in \mathcal{N}$. It is observed that abiding by this constraint does not enforce any other constraints to be violated. A popular way of introducing constraints into the RL problem is by introducing penalties in the reward function \cite{sutton2011reinforcement}. If the constraints are violated, then the agent receives a negative reward. The value of the reward is large enough in a negative direction that cannot be achieved otherwise. There are multiple control elements available at the disposal of transmission control for keeping the grid under the constraints specified in \ref{eq:constraints}, but the Active Voltage Regulator (AVR) attached to each generator is used most frequently to regulate the voltage thresholds in the grid \cite{kundur1994power}. 

\paragraph{Actions} The presented work focuses primarily on IEEE14, but it can be extended to any network which has an AVR attached to each generator. For the RL agent, the voltage reference point of each generator is an element of the action vector $\Act{} = < V_{ref_i} | i \in \mathcal{G} >$. 

By modifying the voltage reference points, the nominal voltage over all the buses is sought. Traditionally, PID controllers are used for such tasks, and they only monitor a single pilot bus. The PID controller tries to keep the voltage of the pilot bus within the prescribed thresholds, but in that effort, it is possible that the voltage on any one of the other 13 buses goes out of the threshold range. The presented work presents a technique by which a model can be trained that monitors the voltages of all the buses and sets the AVR reference points such that the least number of buses go out of the voltage threshold boundary. 

\paragraph{States} As mentioned earlier, the state of the environment is fully observable in the presented case. The learning agent is provided with the voltages of all the buses, i.e.  $\Stat{} = < V_{k} | k \in \mathcal{N} >$, after performing an action. It should be noted that because the agent works in a bandit setting, the \textit{resulting} state, or the \textit{final} state, of the environment does not make any difference on the learning process. The agent only derives the gradient $\PolGradTheta$ with the help of obtained reward, as it is clear from the simplified equation \ref{eq:polGradOne}. The equation is a direct consequence of the proof of policy gradient theorem given in \cite{sutton2011reinforcement}.

\begin{equation}
\label{eq:polGradOne}
\begin{array} {lcl} \PolGradTheta  & = & p(\Stat{1}) \int_{\ActS} \GradTheta{} \PolTheta{} (\Act{}|\Stat{}) \QPi (\Stat{},\Act{}) da  \\ 
& = &  \mathbb{E}_{p(\Stat{1}), a \sim \pi_{\theta}} [\GradTheta{} \log \PolTheta{} (\Act{}|\Stat{}) \QPi{} (\Stat{},\Act{})] \end{array}
\end{equation}

\paragraph{Reward} 

It is clear from the equation \ref{eq:polGradOne} that the value of $\PolGradTheta$ is largely affected by $\QPi (\Stat{},\Act{})$. The value is learned by the \quotes{experience}. The agent performs actions, then based upon the earned reward, the $\QPi$ is learned using the following equation.

\[\QPi(\Stat{t},\Act{t})  =   \mathbb{E}_{\Rew{t},\Stat{t+1} \sim E} [ \Rew{}(\Stat{t},\Act{t}) + \gamma \mathbb{E}_{\Act{t+1} \sim \pi} [\QPi ( \Stat{t+1}, \Act{t+1} ) ]   ] \]

In the presence of a deterministic policy, which is the case in OPF problem, the above equation can be transformed into  equation \ref{eq:qval}, where $\mu : \mathcal{A} \rightarrow \mathcal{S}$ \cite{Lillicrap2016} 

\begin{equation}
\label{eq:qval}
\begin{array} {lcl} 
\QMu(\Stat{t},\Act{t}) & = &  \mathbb{E}_{\Rew{t},\Stat{t+1} \sim E} [ \Rew{}(\Stat{t},\Act{t}) + \QMu ( \Stat{t+1}, \mu(\Stat{t+1}) )  ]
	\end{array}
\end{equation}

It is clear from equations \ref{eq:qval} and \ref{eq:polGradOne} that the reward function plays a critical role in calculating the gradient, which determines the search direction in case of optimization. 

Different reward functions can be devised for the OPF problem, each with the intention of minimizing the distance of the state of the grid from the desired one. Among simpler functions, the squared sum of absolute difference of the resultant state vector $\Stat{f}$ from the nominal state vector, works quite good. As the nominal state vector is $\mathbf{1.0} \in \mathbb{R}^{14}$, so the reward is given by

\begin{equation}
 \Rew{} = -1 \times \left(\sum_{i=1}^{14} |1.0 - \StatEl{f_i}| \rm \right)^2 
\label{eq:reward}
\end{equation}

The reward function is chosen to slow down the momentum when the distance from the desired state is minimum. It also ensures that the gradient near the optimal state does not fluctuate rapidly to take the approximating function quickly far from the local minimum.

\subsubsection{Proposed solution}
\label{sec:sol}

It is proposed that many different RL algorithms can be used in the given situation \cite{arulkumaran2017deep}. The algorithm must be able to work on continuous states. In the recent past, a few algorithms have been proposed for such tasks, for example, Deep Deterministic Policy Gradient(DDPG) algorithm \cite{Lillicrap2016} and Twin Delayed Deep Deterministic policy gradient (TD3) \cite{Fujimoto2018}. In the next section  \ref{sec:casestudies}, results are presented using the aforementioned algorithms. The problem with these algorithms is that they are not parallel algorithms, so the hefty price of environment evaluation makes the optimization procedure slower.  In section \ref{sec:AlgoParallel} a modified version of the DDPG algorithm is discussed that can evaluate the environment in parallel initially, speeding up the process of optimization.

The DDPG algorithm uses the actor-critic method that is based on the Deterministic Policy Gradient (DPG)  \cite{Silver2014} algorithm. The DPG algorithm notes that in equation \ref{eq:qval} the $\QMu$ is only dependent on the environment, so it can be learned by a sub-optimal policy $\beta$. The value function $Q(s,a)$ is coined as \quotes{critic} and is parameterized over $\ThetQ$; $\Critic{}$. The function mapping the policy at a certain time is coined as \quotes{actor} that is also a parameterized function $\Actor{}$. The DDPG algorithm uses Deep Neural Networks (DNN) for the approximation of $\ThetQ$ and $\ThetMu$ by using equation \ref{eq:qmu} over the initial distribution. \cite{Lillicrap2016}. 

\begin{equation}
\label{eq:qmu}
\begin{array} {lcl} \GradTheta{\mu} D  & \approx & \mathbb{E}_{\Stat{t} \sim \beta}  [\GradTheta{\mu} Q(\Stat{},\Act{}|\ThetQ)|_{\Stat{}=\Stat{t},\Act{}={\mu(\Stat{t}|\ThetMu)}} ] \\
 & = & \mathbb{E}_{\Stat{t} \sim \beta}  [\nabla_{\Act{}} Q(\Stat{},\Act{}|\ThetQ)|_{\Stat{}=\Stat{t},\Act{}={\mu(\Stat{t})}} \GradTheta{\mu} \mu(\Stat{t}|\ThetMu) |_{\Stat{}=\Stat{t}} ] 
\end{array}
\end{equation}

The DDPG algorithm uses the \quotes{off-policy} mechanism by slowing down the update of the $\ThetQ$ and $\ThetMu$. It is achieved by creating shadow DNNs. One set of networks gets their weights updated on individual iteration, but the weights of the \quotes{target} network are only updated after multiplying with a delay factor $\tau$. The second algorithm used in the presented work is the Twin Delayed Deep Deterministic policy gradient algorithm (TD3). The TD3 algorithm further enhances the idea of delay and creates slightly different networks for the \quotes{critic}. The details can be seen in \cite{Fujimoto2018}.

Here the environment that receives actions is a simulation component; The Functional Mock-up Unit (FMU). An FMU provides an interface to interact with a simulation component in a transparent way \cite{blochwitz2011functional}. The action values $\Act{t}$ at any instance $t$ are received as inputs to the FMU, the state variables $\Stat{t}$ represent the state of the power grid. According to RL terminology, the FMU here can be considered as the environment $\Env$. The neural network approximating $\ThetMu$ takes initial state $\Stat{1}$ as input and suggests an action $\Act{1}$. When the action values are given to the FMU, the FMU simulates from time $t_0$ to $t_f$ and returns the updated state variables $\Stat{2}$ and the real reward $\Rew{}$. The neural network approximating $\ThetQ$ takes $\Act{1}$ and $\Stat{2}$ as input and assigns a reward of its own $\Rew{}'$. The difference between $\Rew{}$ and $\Rew{}'$ forms the basis of the loss function for both networks. The initial state $\Stat{1}$ is fixed as the power grid is in the nominal state in the beginning. The state $\Stat{2}$ is evaluated over and over again as the result of applying the action $\Act{1}$. The gradient $\GradTheta{\mu}$ is calculated and applied repeatedly until the terminating condition. It is assumed that the FMU produces perfect results of the simulation of the power grid. There is no stochastic error involved. This is equivalent to saying that the environment $\Env$ is deterministic.

\section{Case Studies}
\label{sec:casestudies}

In order to make the experiment reproducible, all the libraries used in the presented work are available open-source. To simulate the power grid scenarios, a Modelica-based library OpenIPSL \cite{baudette2018openipsl} is used. The library is widely acclaimed in the power grids community for its modularity, stability, and robustness. An open-source implementation of Modelica, OpenModelica \cite{openmodelica.org:fritzson:2014}, is used. For solving the system, solvers provided by SUNDIALS \cite{hindmarsh2005sundials} library are used that are wrapped in a Python code, named Assimulo \cite{andersson2015assimulo} and PyFMI \cite{andersson2016pyfmi}.

Two case studies have been chosen to demonstrate the effectiveness of the RL optimization technique. In the first case study, a scenario is modeled in which generator $2$ goes out of order and stops providing any power to the system. In the second case study, both generators $2$ and $8$ go out of order, and voltage reference points of other generators are sought in order to bring each bus in the system within the prescribed voltage threshold.

In both the scenarios presented here, a simulation of the IEEE 14 bus system is used. The system has been used in many studies. The presented work mainly focuses on its 14 buses $B_1, B_2, \dots, B_{14}$ and its five generators $G_1, G_2, G_3, G_6 , G_8$. The numbers given to the generators are indicative of the buses they are attached to $B_1, B_2, B_3, B_6, B_8$. In order to keep the perspective, the action values are also named in the same fashion, i.e. $\Act{} = <a_1, a_2, a_3, a_6, a_8>  $. Each $a_i$ is the reference voltage for the generator $G_i$. 

Certainly, the values of $\Stat{}$ are governed by the changes in $\Act{}$. The noteworthy point in the following discussion is that how each algorithm is able to rapidly take the $\Stat{}$ towards the optimal candidate solution. This is because in comparison to Evolutionary Strategies(ES), here the function $Q$ guides the solution in the right direction \cite{sutton2011reinforcement}.

\subsection{Result interpretation}

\addPlot{0.95}{1/log-ieee_14_82dn-DDPG-s-squaref.pdf}{Figure shows a typical run of one of the proposed algorithms. It shows all the solutions proposed during the run, even the ones that are degenerate solutions. In the future, such degenerate solutions will be removed.}{fig:compfig}

Before looking into the case studies and their result, it is important to mention that it is a simulation-based study, and there are values of $\Act{}$ which lead to degenerate solutions, so the presented results are not plotted with the complete trajectories. The degenerate solutions and insignificant trajectories are omitted from the plots to avoid clutter. For example, as shown in figure \ref{fig:compfig}, the plot is a complete plot including the degenerate solutions and trajectories for all the bus voltages. In order to understand the degenerate solutions, it is important to understand how Modelica works. Input to Modelica is a system of differential equations, which are normally complex enough that they cannot be simulated directly. In order to simplify the system, Modelica applies different simplification algorithms to convert the system to an Ordinary Differential Equation (ODE). Such simplifications impose some constraints which were absent in the original solution. It must be kept in mind that the system already had some constraints even before the simplifications were applied. However the constraints are imposed, the solution may be undefined at some parameter values. In such cases, the values of simulation variables appear to be a constant value. There are other sources of such instabilities, like limitations of the solver. Mostly, a simulation engineer has to accept all these limitations as the area of the system that is undefined. For details on the sources of instabilities and how to tackle them, see \cite{cellier2006continuous}.

In the figure \ref{fig:fulleg}, all the degenerate solutions have been removed, along with the insignificant bus voltage values, yielding a clear graph showing the trend lines. The middle line is the average voltage. A complete description of how the values are acquired for each $B_i$ is coming shortly. Here the intention is to inform that the aforementioned simplification has been applied on the plots to reduce the clutter.

\addTwoPlots{0.49}{1/s-82-DDPG-s-squaref-trend.pdf}{1/s-82-DDPG-s-squaref-action.pdf}{Example of a simplified plot. On the left, the figure shows only the trend all the bus voltages follow, removing all degenerate solutions from figure \ref{fig:compfig}. The solid line plots $\frac{max(\Stat{})+min(\Stat{})}{2}$. The spread around it shows the minimum and maximum values achieved by any bus. The red dots represent the best solution found by the algorithm. On the right are the plots of action values.}{fig:fulleg}

In order to understand the plotted graphs, it is important to understand the simulation process. The simulation component $\SimFMU$ is first provided with the initial values and a simulation time $t_1$. At this moment, no fault is introduced in the system. The simulation component $\SimFMU$ at time $t_1$ presents a healthy state of the power grid. At time $t_1 + \epsilon$ the fault is introduced. Then the system is simulated again till time $t_2$. Certain nominal conditions of the power grid are violated during this simulation. The algorithm proposes a solution in the form of $\Act{}$. The $\Act{}$ is applied to $\SimFMU$ and $\SimFMU$ is simulated till $t_f$, again. The voltages of all the buses at $t_f$ are the state values $\Stat{}$ which are used as observations. In the figure \ref{fig:fulleg}, the left plot shows the values of $\Stat{}$ at $t_f$ for each iteration and the plot on the right shows the values of $\Act{}$ applied at $t_2$. In the middle, there are the trend lines for $\Stat{}$ showing the behavior of the curves.

\subsection{Scenario - 1}
\label{sec:scene-1}
Taking the IEEE 14 bus system as the reference, a scenario is assumed where generator $G_2$ goes out of order and has to be disconnected from the system using a breaker. The system breaks the nominal conditions for the voltage on buses. The algorithm starts its work at that point and finds the optimal values of $\Act{}$ that takes the system back into the nominal condition, or as close as possible. When the generator $G_2$ is disconnected, there is no purpose in applying $a_2$. The action vector is reduced to $\Act{} = <a_1, a_3, a_6, a_8>$.

\addTwoPlots{0.45}{2/s-1-DDPG-s-sqaurf-trend.pdf}{2/s-1-DDPG-s-sqaurf-action.pdf}{DDPG Algorithm (scenario-1): on the left are the trend lines for bus voltages, on the right is the evolution of action values. The red dots show where the optimal results are found.}{fig:ddpg-2}

\addTwoPlots{0.45}{3/s-1-TD3-s-sqaurf-trend.pdf}{3/s-1-TD3-s-sqaurf-action.pdf}{TD3 algorithm (scenario-1): on the left are trend lines for bus voltages, on the right is the evolution of action values. There is no red dot here because all proposed solutions required more than one transformer action.}{fig:td3-2}

\addTwoPlots{0.45}{2/V-99.pdf}{3/V-95.pdf}{These are the best solutions found by TD3 and DDPG algorithms for scenario-1. Left one is the solution found by the DDPG algorithm, on the right is the solution found by the TD3 algorithm.}{fig:best-sol}
\afterpage{\clearpage}

As can be seen in the figure \ref{fig:best-sol}, the DDPG algorithm is able to achieve very good results. Only $B_8$ is above the desired threshold of $1.05$. However, the TD3 algorithm is not able to produce a very good result, as shown on the right side of the figure \ref{fig:best-sol}. Here two buses $B_1$ and $B_2$ are above the nominal threshold $1.05$. Secondly, it can be seen that the TD3 algorithm does not offer any advantage in this case. Rather, the results are worse than the DDPG algorithm. The reason being, TD3 algorithm is designed to tackle the residue that accumulates at each evaluation of the value function, when the trajectory of actions is considerably long. In the presented case, there is virtually no trajectory of actions at all. Because it is a bandit situation as described in the section \ref{sec:problem}. Moreover, the TD3 algorithm provides a policy smoothing mechanism, which in this case is counterproductive. It dampens the response of the neural networks, so they do not adapt swiftly, giving smoother action trajectories as compared to the DDPG algorithm.  
\subsection{Scenario - 2}
\label{sec:scene-2}

Similar to scenario-1 mentioned earlier, here it is assumed that the $G_2$ and $G_8$ have both gone out of order. In such a situation, the voltage drop is experienced on almost all the buses. In this scenario, the action vector is $\Act{} = <a_1, a_3, a_6>$. As can be seen in figure \ref{fig:82-best-sol} that both DDPG and TD3 algorithms were able to find a considerably good solution. The figure on the left side of figure \ref{fig:82-best-sol} shows that only $B_1$ is greater than the nominal threshold of $1.05$, but that can be controlled using a transformer. The right side of the figure \ref{fig:82-best-sol} shows that again $B_1$ is greater than the nominal threshold $1.05$, while $B_2$ is also a little over the line.

\addTwoPlots{0.49}{4/V-88.pdf}{5/V-99.pdf}{These are the best solutions found by both DDPG and TD3 algorithms for scenario-2. On the left side is the result found by the DDPG algorithm and on the right side is the results found by the TD3 algorithm. Here both generators $2$ and $8$ were put offline.}{fig:82-best-sol}

\addTwoPlots{0.49}{4/s-82-DDPG-s-squaref-trend.pdf}{4/s-82-DDPG-s-squaref-action.pdf}{DDPG Algorithm (scenario-2): on the left are the trend lines for bus voltages, and on the right is the evolution of action values.}{fig:ddpg-82}

\paragraph{Adding Transformer Actions}
In the solutions proposed above, mostly only one bus is observed to break the $1.05$ threshold. This problem can be mitigated using the transformer operation. It is much cheaper and efficient to perform an operation over one bus in comparison to performing different operations on many buses. Secondly, if the optimal values and required transformer operations for such scenarios are cached, then mitigating such faults may become automatic.

\addTwoPlots{0.49}{5/s-82-TD3-s-squarf-trend.pdf}{5/s-82-TD3-s-squarf-action.pdf}{TD3 algorithm (scenario-2): on the left are trend lines for bus voltages, and on the right is the evolution of action values.}{fig:td3-82}
\afterpage{\clearpage}

\section{Limitations}
\label{sec:limits}

As described in \cite{Mnih2015}, RL is unstable and sometimes divergent when a nonlinear component, such as a neural network, is used for approximation of $Q$ value. Although using delayed update of the network \cite{Lillicrap2016} and replay buffer technique mentioned in  \cite{Mnih2015} can control the instabilities by reducing the correlation in the experiences, but they are not proven to make the system convergent. The same divergence can be seen here. With the number of iterations, the solution should have converged to the correct solution. Instead, it converges to the minimum action values, increasing the distance from the global optimum. Following are two reasons for this issue
\begin{enumerate}
\item Exploration noise: The exploration noise allows the algorithm to explore all possible action values $\Act{}$, based upon the feedback of the $\Env$ the neural network learns how to modify the $\Act{}$ values. Equation \ref{eq:performObj} determines the change in value. If the variance in the normal exploration noise is too low, then there is a risk of missing the values leading to the global optimum. If the variance is taken too high, then the experience correlation can take the future $Q$ values in the wrong direction. 
\item Experience Correlation: looking at the equation \ref{eq:polGradOne} the policy gradient is completely dependent upon the previous experience. The correlation among the past experiences is beneficial as it drives the solution in the right direction, but it is also problematic when the algorithm reaches closer to the optimal value. The bias in the experience data set imposes an implicit \textit{momentum} that leads the average gradient value away from the current gradient. This is especially problematic when an abrupt change in the gradient is required for convergence.
\end{enumerate}

Taking into consideration the above limitations, in the following sections, two solutions are proposed. One solution finds the near-optimal solution quite swiftly, while the other solution deals with the problem of divergence. Nevertheless, these solutions are domain-specific and specifically designed for the problem at hand.

\section{Improvement 1: Parallelism}
\label{sec:AlgoParallel}

As mentioned in section \ref{sec:sol}, the DDPG algorithm uses the difference between the reward given by the actual environment and the reward predicted by the critic to predict the response from the environment accurately. In the problem under discussion, the reward function is a simple aggregate over the state variables. So in order to improve the convergence rate, the algorithm can be modified to run in two stages. In the first stage, the critic is approximated independently by running many parallel calls to the environment and reducing the mean squared distance between the predicted states and the actual states. In the next stage, the normal policy iteration is executed to find the optimal solution parameters. 

In other words the critic $Q(\Stat{}, \Act{}|\ThetQ)$ is learned by a network, say $\Phi_q$. The network $\Phi_q$ has weights associated with each layer numbered through $1,2,3,\dots, n$. The respective weights for each layer can be named as $w_1, w_2, w_3, \dots, w_n$. The idea is to utilize the fact that the environment in the presented case is a dynamic system, and because the reward function is a simple squared sum of absolute differences, so a neural network representing the environment will not be much different from $\Phi_q$. Let us call the network representing the environment as $\overline{\Phi_q}$, where $\overline{\Phi_q}$ has $n-1$ layers as compared $\Phi_q$ which has $n$ layers. The environment here is a simulation instantiated using an FMU that can be symbolized as $\SimFMU$. This means that we can assume that $\Phi_q \approx w_n \ast \overline{\Phi_q}$, where $\ast$ is an arbitrary tensor operation defined by the structure of the neural network. Training  $\overline{\Phi_q}$ such that $\overline{\Phi_q} \approx \SimFMU$ will lead to a better approximation  $\Phi_q \approx Q(\Stat{}, \Act{}|\ThetQ)$, because $Q(\Stat{}, \Act{}|\ThetQ)$ is predicated upon $\SimFMU$.

\begin{algorithm}[h]
\begin{algorithmic}[1]
\State Randomly initialize the weights $\overline{\ThetQ}$ of $\overline{\Phi_q}$
\State Initialize $\Act{1}$ by already prescribed nominal values
\State Initialize $t$ by an appropriate simulation time
\For {$i \gets 1 \to M$} 
	\State Initialize $\Stat{1} \gets  \textsc{Initialize\_FMU}(\Act{1},t)$
	\State select $\Act{i}$ randomly from a uniform distribution
	\State $\Stat{i} \gets \SimFMU(\Act{i}, t+\delta)$
	\State $\overline{\Stat{i}} \gets \overline{\Phi_q}(\Act{i})$
	\State Update $\overline{\Phi_q}$ by minimizing the loss: $L = \norm{\Stat{i} - \overline{\Stat{i}}}_2$
\EndFor
\State  Initialize first $n-1$ layers of critic network $\Phi_q$ using $\overline{\ThetQ}$ 
\State Proceed DDPG algorithm \cite{Lillicrap2016}, with initializing only $n^{th}$ layer of $\ThetQ$ 
\Procedure {Initialize\_FMU} {$\Act{}$, $t$}
	\State Initialize $\SimFMU$ according to the FMI standard
	\State Provide $\Act{}$ to $\SimFMU$ as arguments and simulate till $t$,  $\Stat{} \gets \SimFMU(\Act{}, t)$ 
	\State \Return the observation variables $\Stat{}$ 
\EndProcedure
\caption{Faster Training}
\end{algorithmic}
\end{algorithm}

\addTwoPlots{0.49}{6/V-239.pdf}{7/V-217.pdf}{Parallel DDPG Algorithm (scenario 1 and 2): on the left is the best solution for scenario-1. On the right is the best solution for scenario-2.}{fig:pddpg-best}

\addTwoPlots{0.49}{6/s-1-DDPG-p-squaref-trend.pdf}{6/s-1-DDPG-p-squaref-action.pdf}{Parallel DDPG Algorithm (scenario-1): on the left are trend lines for bus voltages, and on the right is the evolution of action values. The red dots show the positions of the solutions.}{fig:pddpg-2}

\addTwoPlots{0.49}{7/s-82-DDPG-p-squaref-trend.pdf}{7/s-82-DDPG-p-squaref-action.pdf}{Parallel DDPG Algorithm (scenario-2): on the left are trend lines for bus voltages, and on the right is the evolution of action values.}{fig:pddpg-82}

In figures \ref{fig:pddpg-2} and \ref{fig:pddpg-82}, it can be seen that the algorithm is able to find a good solution only after $89$ and  $67$ iterations, respectively, neglecting the initial $150$ evaluations. This is approximately $10\%$ faster in the case of scenario-1 and  $25\%$ faster for scenario-2. The solution is not only found sooner, the quality of the solution is also much better than all previously proposed solutions. All bus voltages are close to each other and within the desired range, with the exception of only one bus. The important thing to note is that the first $150$ evaluations were executed in parallel, so the time spent on these evaluations could be equal to a single evaluation of other solutions if there are $150$ processing cores available. In the experiment presented here, $30$ cores were used in parallel.   

\section{Improvement 2: Reward Function}
\label{sec:rewardFunc}
Close examination of the problem reveals certain patterns. Utilizing these can lead to design a reward function that is more stable. Following are some results that can be concluded after close inspection of the solution trajectories in the figures above.
\begin{itemize}
\item The reward function in equation \ref{eq:reward} works well at the beginning when the average bus voltage is greater than the nominal threshold. 
\item When the average bus voltage is within the nominal threshold, the search should remain there and keep finding better solutions.
\item When the average bus voltage is below the nominal threshold, then the direction of the gradient should change so the search can continue within the nominal boundaries.
\end{itemize}

According to the above considerations a piece-wise function is necessary. In the reward function given below $v = \frac{max(\Stat{}) + min(\Stat{})}{2}$ and $C$ is the maximum deviation a stable solution can have. In other words, it can be the maximum value $\hat{r} = \sum_{i=1}^{14} |1.0 - \StatEl{f_i}|$ can ever achieve. The last term in the equation \ref{eq:preward} ensures that the search direction changes to lead the search towards the optimal threshold value.

\begin{equation}
r = 
\begin{cases} 
      -1 \times \hat{r} & v > 1.05 \\
      -1 \times (max(\Stat{}) - min(\Stat{}))^2 & 0.95\leq v \leq 1.05 \\
      C - max(|\mathbf{1.0} - \Stat{}|)& v < 0.95  
   \end{cases}
  \label{eq:preward}
\end{equation}

\addTwoPlots{0.45}{8/s-1-DDPG-p-reward-trend.pdf}{8/s-1-DDPG-p-reward-action.pdf}{ Voltage and action trajectories of scenario-1, where $G_2$ goes offline. Here the piece-wise reward function was used.}{fig:prew-2}

\addTwoPlots{0.45}{9/s-82-DDPG-p-reward-trend.pdf}{9/s-82-DDPG-p-reward-action.pdf}{ Voltage and action trajectories of scenario-2, where $G_2$ and $G_8$ go offline. Here the piece-wise reward function was used.}{fig:prew-82}

The second important aspect is the correlation of previous experiences. In order to break the correlation, the replay buffer has to be reinitialized at appropriate times. Again based upon the domain-specific knowledge, the replay buffer is reinitialized when $v$ shifts from one bracket to the other. Figures \ref{fig:prew-2} and \ref{fig:prew-82} show two example runs of search algorithm for the both scenarios. It can be seen that the reward function enables the search algorithm to look for the best solutions in the ripest area for optimal solutions.

\section{Discussion}
\label{sec:discussion}
Although the goal of the research was to find a way to propose optimal values of generators in case of a failure, some interesting aspects have emerged from the study that are discussed as follows.
 
\subsection{River run analogy}
The idea of the reward function given in \ref{eq:preward} can be understood using an analogy. Suppose there is new land found by some farmers. They want to cultivate it in the most productive way possible, where the fruit is brought with the least effort. However, the terrain is such that the river does not pass through the most propitious of areas. They do have limited resources to diverge the direction of the river and to direct it towards the favorable lands, but the resources are not enough to change the direction of the river completely at will. Also, their knowledge about where the most favorable areas lie is not accurate. In such a situation, the best they can do is to put hurdles in the way of the river to redirect it towards favorable areas. Although the river always follows the terrain so it always diverges towards its natural stream, restricting its movement will enable the farmers to cultivate more favorable lands.

The same is the situation here. An optimization problem is a search problem. In heuristic optimization techniques, different heuristics are used to direct the search towards more ripe areas for optimal solutions. For example, evolutionary strategies try to use natural selection to get to the optimal solution after certain generations. In a way, it is also a method of stopping the river from following the terrain and changing its direction towards a heuristically optimal path. In the presented technique, the $Q$ value is the force of the river, which can be directed towards favorable lands using the reward function by creating suitable hurdles in the reward function. This is left to be proven that such an effort will surely be able to find the optimal solution or not. Although for Q-learning, it is proven that enough iterations will lead to the optimal solution, for that, there is no divergence problem as mentioned in \cite{Mnih2015} and in section \ref{sec:limits}.

\subsection{The optimal solution}
It is clear from the above discussion that the proposed algorithm could not find an ideal solution; A reference voltage of all the working generators that can take the bus voltages within the nominal threshold. It is deduced based upon the experiments given above that either there is no such value in scenarios $1$ and $2$ which can take the system to optimal condition, or the path toward it is littered with unstable solutions. If we assume that the real reason is the second case, the likelihood of getting to the optimal solution depends upon the position of the optimal solution within the unstable solutions. If the optimal solution is completely surrounded by unstable solutions, then it is highly unlikely for the proposed algorithm that the search will ever reach there. For example, in the figure \ref{fig:unstab} a near-optimal solution for scenario-1 is shown. It can be seen the voltages on each bus are not stable. 

\addTwoPlots{0.49}{10/V-193.pdf}{10/angle-193.pdf}{ On the left is the bus voltages, on the right is the phase angle. The difference in phase angle shows that the generators are not synchronized, so the solution is not a stable and valid solution.}{fig:unstab}

One can assume that if these types of solutions are surrounding the optimal solution, then reaching the optimal solution will be difficult as such unstable solutions take far more time to solve. Such solutions appear more often in scenario-1, costing a huge computation time for a single evaluation increasing the complete execution time tremendously. That is why figure \ref{fig:prew-2} have far less evaluations than figure \ref{fig:prew-82}. 

\section{Conclusion}
\label{sec:conc}
In the presented article, a novel technique is discussed to find stable optimal values of reference voltages in case of generator failure. The technique is based upon reinforcement learning. Two scenarios from IEEE 14 bus system have been chosen. In the first scenario, generator $2$ fails, while in the second scenario, generators $2$ and $8$ fail. Although the algorithm was not able to find the ideal solution---the reference voltages that take the system back to nominal condition---the algorithm was able to find numerous solutions with only one bus voltage out of the nominal range. The study has opened up new questions for research. For example, to find a reward function that can take the search to the optimal solution swiftly. One main issue that has been identified in the presented technique is that it cannot be completely made parallel. An algorithm is proposed that improves the performance by introducing an initialization phase that is parallel and enables the algorithm to reach the solution swiftly. Still, it is not completely parallel. In the future, these questions will be addressed.

\bibliographystyle{unsrtnat}
\bibliography{RLOptGrid.bib}

\begin{thebibliography}{31}
\providecommand{\natexlab}[1]{#1}
\providecommand{\url}[1]{\texttt{#1}}
\expandafter\ifx\csname urlstyle\endcsname\relax
  \providecommand{\doi}[1]{doi: #1}\else
  \providecommand{\doi}{doi: \begingroup \urlstyle{rm}\Url}\fi

\bibitem[Martins et~al.(2019)Martins, Felgueiras, Smitkova, and
  Caetano]{martins2019analysis}
Florinda Martins, Carlos Felgueiras, Miroslava Smitkova, and N{\'\i}dia
  Caetano.
\newblock Analysis of fossil fuel energy consumption and environmental impacts
  in european countries.
\newblock \emph{Energies}, 12\penalty0 (6):\penalty0 964, 2019.

\bibitem[Gu{\'e}rin et~al.(2011)Gu{\'e}rin, Lefebvre, Mboup, Par{\'e}d{\'e},
  Lemains, and Ndiaye]{guerin2011hybrid}
Fran{\c{c}}ois Gu{\'e}rin, Dimitri Lefebvre, Alioune~Badara Mboup, Jean-Yves
  Par{\'e}d{\'e}, Eric Lemains, and Pape Alioune~Sarr Ndiaye.
\newblock Hybrid modeling for performance evaluation of multisource renewable
  energy systems.
\newblock \emph{IEEE transactions on automation science and engineering},
  8\penalty0 (3):\penalty0 570--580, 2011.

\bibitem[Ernst et~al.(2004)Ernst, Glavic, and Wehenkel]{ernst2004power}
Damien Ernst, Mevludin Glavic, and Louis Wehenkel.
\newblock Power systems stability control: reinforcement learning framework.
\newblock \emph{IEEE transactions on power systems}, 19\penalty0 (1):\penalty0
  427--435, 2004.

\bibitem[Dimeas and Hatziargyriou(2010)]{dimeas2010multi}
AL~Dimeas and ND~Hatziargyriou.
\newblock Multi-agent reinforcement learning for microgrids.
\newblock In \emph{IEEE PES General Meeting}, pages 1--8. IEEE, 2010.

\bibitem[Li et~al.(2012)Li, Wu, He, and Chen]{li2012optimal}
Fu-Dong Li, Min Wu, Yong He, and Xin Chen.
\newblock Optimal control in microgrid using multi-agent reinforcement
  learning.
\newblock \emph{ISA transactions}, 51\penalty0 (6):\penalty0 743--751, 2012.

\bibitem[Diao et~al.(2019)Diao, Wang, Shi, Chang, Duan, and
  Zhang]{diao2019autonomous}
Ruisheng Diao, Zhiwei Wang, Di~Shi, Qianyun Chang, Jiajun Duan, and Xiaohu
  Zhang.
\newblock Autonomous voltage control for grid operation using deep
  reinforcement learning.
\newblock In \emph{2019 IEEE Power \& Energy Society General Meeting (PESGM)},
  pages 1--5. IEEE, 2019.

\bibitem[Huang et~al.(2019)Huang, Huang, Hao, Tan, Fan, and
  Huang]{huang2019adaptive}
Qiuhua Huang, Renke Huang, Weituo Hao, Jie Tan, Rui Fan, and Zhenyu Huang.
\newblock Adaptive power system emergency control using deep reinforcement
  learning.
\newblock \emph{IEEE Transactions on Smart Grid}, 11\penalty0 (2):\penalty0
  1171--1182, 2019.

\bibitem[Yang et~al.(2019)Yang, Wang, Sadeghi, Giannakis, and Sun]{yang2019two}
Qiuling Yang, Gang Wang, Alireza Sadeghi, Georgios~B Giannakis, and Jian Sun.
\newblock Two-timescale voltage control in distribution grids using deep
  reinforcement learning.
\newblock \emph{IEEE Transactions on Smart Grid}, 11\penalty0 (3):\penalty0
  2313--2323, 2019.

\bibitem[Hua et~al.(2019)Hua, Qin, Hao, and Cao]{hua2019optimal}
Haochen Hua, Yuchao Qin, Chuantong Hao, and Junwei Cao.
\newblock Optimal energy management strategies for energy internet via deep
  reinforcement learning approach.
\newblock \emph{Applied Energy}, 239:\penalty0 598--609, 2019.

\bibitem[Wang et~al.(2019)Wang, Yu, Gao, and Shi]{wang2019safe}
Wei Wang, Nanpeng Yu, Yuanqi Gao, and Jie Shi.
\newblock Safe off-policy deep reinforcement learning algorithm for volt-var
  control in power distribution systems.
\newblock \emph{IEEE Transactions on Smart Grid}, 2019.

\bibitem[Wu et~al.(2020)Wu, Wei, Liu, Quan, and Li]{wu2020battery}
Jingda Wu, Zhongbao Wei, Kailong Liu, Zhongyi Quan, and Yunwei Li.
\newblock Battery-involved energy management for hybrid electric bus based on
  expert-assistance deep deterministic policy gradient algorithm.
\newblock \emph{IEEE Transactions on Vehicular Technology}, 2020.

\bibitem[Marot et~al.(2020)Marot, Donnot, Romero, Donon, Lerousseau,
  Veyrin-Forrer, and Guyon]{marot2020learning}
Antoine Marot, Benjamin Donnot, Camilo Romero, Balthazar Donon, Marvin
  Lerousseau, Luca Veyrin-Forrer, and Isabelle Guyon.
\newblock Learning to run a power network challenge for training topology
  controllers.
\newblock \emph{Electric Power Systems Research}, 189:\penalty0 106635, 2020.

\bibitem[Baudette et~al.(2018)Baudette, Castro, Rabuzin, Lavenius, Bogodorova,
  and Vanfretti]{baudette2018openipsl}
Maxime Baudette, Marcelo Castro, Tin Rabuzin, Jan Lavenius, Tetiana Bogodorova,
  and Luigi Vanfretti.
\newblock Openipsl: Open-instance power system library update 1.5 to "itesla
  power systems library (ipsl): A modelica library for phasor time-domain
  simulations".
\newblock \emph{SoftwareX}, 7:\penalty0 34--36, 2018.

\bibitem[Mohagheghi et~al.(2018)Mohagheghi, Alramlawi, Gabash, and
  Li]{mohagheghi2018survey}
Erfan Mohagheghi, Mansour Alramlawi, Aouss Gabash, and Pu~Li.
\newblock A survey of real-time optimal power flow.
\newblock \emph{Energies}, 11\penalty0 (11):\penalty0 3142, 2018.

\bibitem[Lavaei and Low(2011)]{lavaei2011zero}
Javad Lavaei and Steven~H Low.
\newblock Zero duality gap in optimal power flow problem.
\newblock \emph{IEEE Transactions on Power Systems}, 27\penalty0 (1):\penalty0
  92--107, 2011.

\bibitem[Wood et~al.(2013)Wood, Wollenberg, and Shebl{\'e}]{wood2013power}
Allen~J Wood, Bruce~F Wollenberg, and Gerald~B Shebl{\'e}.
\newblock \emph{Power generation, operation, and control}.
\newblock John Wiley \& Sons, 2013.

\bibitem[Lin et~al.(2018)Lin, Ju, Yong-hua, Qi, and Wang]{lin2018novel}
Yi~Lin, Yun-tao Ju, Z~Yong-hua, Zhi-nan Qi, and Jing Wang.
\newblock A novel sequence-phase coupled model for active distribution network
  based on modelica.
\newblock In \emph{2018 2nd IEEE Conference on Energy Internet and Energy
  System Integration (EI2)}, pages 1--5. IEEE, 2018.

\bibitem[Silver et~al.(2014)Silver, Lever, Heess, Degris, Wierstra, and
  Riedmiller]{Silver2014}
David Silver, Guy Lever, Nicolas Heess, Thomas Degris, Daan Wierstra, and
  Martin Riedmiller.
\newblock {Deterministic policy gradient algorithms}.
\newblock In \emph{31st International Conference on Machine Learning, ICML
  2014}, volume~1, pages 605--619, 2014.
\newblock ISBN 9781634393973.

\bibitem[Sutton et~al.(2000)Sutton, McAllester, Singh, and
  Mansour]{sutton2000policy}
Richard~S Sutton, David~A McAllester, Satinder~P Singh, and Yishay Mansour.
\newblock Policy gradient methods for reinforcement learning with function
  approximation.
\newblock In \emph{Advances in neural information processing systems}, pages
  1057--1063, 2000.

\bibitem[Sutton and Barto(2011)]{sutton2011reinforcement}
Richard~S Sutton and Andrew~G Barto.
\newblock Reinforcement learning: An introduction.
\newblock 2011.

\bibitem[Kundur et~al.(1994)Kundur, Balu, and Lauby]{kundur1994power}
Prabha Kundur, Neal~J Balu, and Mark~G Lauby.
\newblock \emph{Power system stability and control}, volume~7.
\newblock McGraw-hill New York, 1994.

\bibitem[Lillicrap et~al.(2016)Lillicrap, Hunt, Pritzel, Heess, Erez, Tassa,
  Silver, and Wierstra]{Lillicrap2016}
Timothy~P. Lillicrap, Jonathan~J. Hunt, Alexander Pritzel, Nicolas Heess, Tom
  Erez, Yuval Tassa, David Silver, and Daan Wierstra.
\newblock {Continuous control with deep reinforcement learning}.
\newblock In \emph{4th International Conference on Learning Representations,
  ICLR 2016 - Conference Track Proceedings}, 2016.

\bibitem[Arulkumaran et~al.(2017)Arulkumaran, Deisenroth, Brundage, and
  Bharath]{arulkumaran2017deep}
Kai Arulkumaran, Marc~Peter Deisenroth, Miles Brundage, and Anil~Anthony
  Bharath.
\newblock Deep reinforcement learning: A brief survey.
\newblock \emph{IEEE Signal Processing Magazine}, 34\penalty0 (6):\penalty0
  26--38, 2017.

\bibitem[Fujimoto et~al.(2018)Fujimoto, {Van Hoof}, and Meger]{Fujimoto2018}
Scott Fujimoto, Herke {Van Hoof}, and David Meger.
\newblock {Addressing Function Approximation Error in Actor-Critic Methods}.
\newblock \emph{35th International Conference on Machine Learning, ICML 2018},
  4:\penalty0 2587--2601, 2018.

\bibitem[Blochwitz et~al.(2011)Blochwitz, Otter, Arnold, Bausch, Clauss,
  Elmqvist, Junghanns, Mauss, Monteiro, Neidhold,
  et~al.]{blochwitz2011functional}
Torsten Blochwitz, Martin Otter, Martin Arnold, Constanze Bausch, Christoph
  Clauss, Hilding Elmqvist, Andreas Junghanns, Jakob Mauss, Manuel Monteiro,
  Thomas Neidhold, et~al.
\newblock The functional mockup interface for tool independent exchange of
  simulation models.
\newblock In \emph{Proceedings of the 8th International Modelica Conference},
  pages 105--114. Link{\"o}ping University Press, 2011.

\bibitem[Fritzson(2015)]{openmodelica.org:fritzson:2014}
Peter Fritzson.
\newblock \emph{Principles of Object-Oriented Modeling and Simulation with
  Modelica 3.3: A Cyber-Physical Approach}.
\newblock Wiley-IEEE Press, 2 edition, April 2015.
\newblock ISBN 978-1-118-85912-4.

\bibitem[Hindmarsh et~al.(2005)Hindmarsh, Brown, Grant, Lee, Serban, Shumaker,
  and Woodward]{hindmarsh2005sundials}
Alan~C Hindmarsh, Peter~N Brown, Keith~E Grant, Steven~L Lee, Radu Serban,
  Dan~E Shumaker, and Carol~S Woodward.
\newblock Sundials: Suite of nonlinear and differential/algebraic equation
  solvers.
\newblock \emph{ACM Transactions on Mathematical Software (TOMS)}, 31\penalty0
  (3):\penalty0 363--396, 2005.

\bibitem[Andersson et~al.(2015)Andersson, F{\"u}hrer, and
  {\AA}kesson]{andersson2015assimulo}
Christian Andersson, Claus F{\"u}hrer, and Johan {\AA}kesson.
\newblock Assimulo: A unified framework for ode solvers.
\newblock \emph{Mathematics and Computers in Simulation}, 116:\penalty0 26--43,
  2015.

\bibitem[Andersson et~al.(2016)Andersson, {\AA}kesson, and
  F{\"u}hrer]{andersson2016pyfmi}
Christian Andersson, Johan {\AA}kesson, and Claus F{\"u}hrer.
\newblock \emph{Pyfmi: A python package for simulation of coupled dynamic
  models with the functional mock-up interface}.
\newblock Centre for Mathematical Sciences, Lund University Lund, 2016.

\bibitem[Cellier and Kofman(2006)]{cellier2006continuous}
Fran{\c{c}}ois~E Cellier and Ernesto Kofman.
\newblock \emph{Continuous system simulation}.
\newblock Springer Science \& Business Media, 2006.

\bibitem[Mnih et~al.(2015)Mnih, Kavukcuoglu, Silver, Rusu, Veness, Bellemare,
  Graves, Riedmiller, Fidjeland, Ostrovski, Petersen, Beattie, Sadik,
  Antonoglou, King, Kumaran, Wierstra, Legg, and Hassabis]{Mnih2015}
Volodymyr Mnih, Koray Kavukcuoglu, David Silver, Andrei~A. Rusu, Joel Veness,
  Marc~G. Bellemare, Alex Graves, Martin Riedmiller, Andreas~K. Fidjeland,
  Georg Ostrovski, Stig Petersen, Charles Beattie, Amir Sadik, Ioannis
  Antonoglou, Helen King, Dharshan Kumaran, Daan Wierstra, Shane Legg, and
  Demis Hassabis.
\newblock {Human-level control through deep reinforcement learning}.
\newblock \emph{Nature}, 518\penalty0 (7540):\penalty0 529--533, 2015.
\newblock ISSN 14764687.
\newblock \doi{10.1038/nature14236}.

\end{thebibliography}

\end{document}